%% file: neurips_2026.tex
\documentclass{article}

\usepackage[main, final]{neurips_2026}
\usepackage{subfig}

\usepackage[utf8]{inputenc} 
\usepackage[T1]{fontenc}    
\usepackage{hyperref}       
\usepackage{url}            
\usepackage{booktabs}       
\usepackage{amsfonts}       
\usepackage{nicefrac}       
\usepackage{microtype}      
\usepackage{xcolor}         
\usepackage[dvipsnames, svgnames, x11names,table,xcdraw]{xcolor}
\definecolor{HeaderGray}{HTML}{D0CECE}
\definecolor{Stripe}{HTML}{E7E6E6}
\definecolor{OursBg}{HTML}{FFF2CC}
\usepackage{amsmath}
\usepackage{algorithm}
\usepackage{algpseudocode}
\usepackage{graphicx}
\usepackage{enumitem}
\usepackage{multirow}
\usepackage{wrapfig}
\title{DetectViT: Test-time Backdoor Detection for Vision Transformers via Inter-Head Attention Discrepancy}

\author{%
  Siquan Huang$^{1}$ \quad
  Yijiang Li$^{2}$ \quad
  Xingfu Yan$^{3,*}$ \quad
  Ningzhi Gao$^{1}$ \quad
  Leyu Shi$^{1}$ \quad
  Ying Gao$^{1,}$\thanks{Corresponding authors: Ying Gao and Xingfu Yan.} \\[4pt]
  {$^{1}$South China University of Technology \quad
   $^{2}$University of California, San Diego} \\[4pt]
  {$^{3}$South China Normal University} \\[4pt]
  {\texttt{\{cssiquanhuang, csgaonz, csshileyu\}@mail.scut.edu.cn}, \
   \texttt{gaoying@scut.edu.cn}} \\[4pt]
  {\texttt{yijiangli@ucsd.edu}, \
   \texttt{xfyan78@163.com}}
}

\begin{document}

\maketitle

\begin{abstract}
  Vision Transformers (ViTs) now underpin the visual encoders of most multimodal models, but this success comes at a cost: relying on third-party pretrained checkpoints leaves these systems open to backdoor attacks. Existing test-time detection methods rely on either the predicted label or intermediate embeddings, limiting their applicability across diverse tasks and incurring substantial computational overhead.
  In this work, we study how backdoor triggers affect the multi-head attention mechanism of ViTs and identify two distinct anomalous patterns: \textit{complete attention hijacking}, causing the attention to exhibit abnormally high inter-head spatial overlap, and \textit{partial attention hijacking}, causing an abnormally large variance in the concentration of the attention distribution. Motivated by these findings, we propose \textbf{DetectViT}, a test-time backdoor detection method that requires neither training data, model outputs, nor any prior knowledge of the trigger. DetectViT quantifies the two hijacking phenomena via an inter-head consistency score and an inter-head entropy variance score, with thresholds estimated from a small out-of-distribution (OOD) calibration set drawn from an arbitrary source. Across four representative attacks, diverse architectures (DeiT, CLIP, and LLaVA), and downstream tasks (image classification and captioning), DetectViT substantially outperforms all baselines with as few as 64--128 OOD calibration samples, achieving a true positive rate of $100\%$ and a false positive rate of $0.17\%$ on BadCLIP. Notably, it introduces negligible additional overhead by relying solely on attention weights already computed during inference. Code is available at: \url{https://github.com/siquanhuang/DetectViT}.
\end{abstract}

\section{Introduction}
Vision Transformers (ViTs)~\cite{dosovitskiy2020image, pmlr-v139-touvron21a, mehta2021mobilevit} have emerged predominantly as the backbone for visual representation learning, serving as the de facto visual encoder in multimodal models~\cite{radford2021learning, mustafa2022multimodal} and large vision-language models (LVLMs)~\cite{zhu2023minigpt, liu2024improved, bai2023qwen, yang2025qwen3}. However, training ViTs from scratch has become increasingly prohibitive as model scales continue to grow~\cite{zhai2022scaling, dehghani2023scaling}. Practitioners therefore rely on third-party pretrained checkpoints, exposing their systems to backdoor attacks~\cite{chen2017targeted, Huang_2023_ICCV,huang2025scope,huang2025fedid, gu2017badnets}. Specifically, an adversary can take advantage of this by injecting a backdoor directly into the pretrained model.
This threat is further amplified by ViT-specific attacks that exploit the self-attention mechanism to achieve high attack success while remaining stealthy~\cite{lv2021dbia, yuan2023you, zheng2023trojvit, wang2025attention}.

Existing backdoor detection methods commonly require access to training data to estimate a reference distribution, enabling effective filtering of poisoned samples during the training phase~\cite{pan2023asset, rong2025backdoor, yuanactivation}. Yet, this assumption is often unrealistic when practitioners obtain pretrained ViTs from third-party sources with no knowledge of the original training process. Moreover, backdoors embedded in pretrained checkpoints can survive downstream fine-tuning~\cite{jia2022badencoder, tao2024distribution, liang2024badclip, liu2025stealthy, yuan2023you, zheng2023trojvit}, underscoring the need for test-time detection.
To address this threat, existing test-time methods predominantly inspect the model's output on suspicious samples, either through predicted labels or intermediate embeddings. However, both paradigms suffer from fundamental limitations. Label-based methods~\cite{gao2019strip, doan2023defending, guoscale, liu2023detecting} are inherently restricted to classification tasks and cannot handle open-ended outputs such as image captioning or visual question answering, where no discrete label space exists. Embedding-based methods~\cite{hou2025dede, huang2026backdoorids} sidestep this restriction but are blind to attacks that target the classification head rather than the encoder~\cite{zheng2023trojvit}, leaving such backdoors entirely undetected. Moreover, both paradigms typically rely on generating multiple augmented copies of each test sample \cite{liu2023detecting,guoscale,doan2023defending,huang2026backdoorids,songtest}, introducing substantial inference-time overhead. These limitations motivate a fundamentally different perspective: \textit{Can we detect backdoor samples by directly probing the internal computation of the ViT itself, without any training data or model outputs?}


To address this question, we turn to the core component of ViTs, the multi-head attention mechanism, and ask: \textit{What does a backdoor trigger do to attention?} While prior work~\cite{yuan2023you,zheng2023trojvit,rong2025backdoor} has shown that triggers can hijack heads' attention toward themselves, we go further by analyzing this hijacking through an \emph{inter-head} lens. Specifically, on clean images, different attention heads naturally attend to spatially distinct regions through their separate representational subspaces, and exhibit broadly comparable concentration across heads, owing to their shared architecture and optimization objective. On backdoored images, however, we observe that triggers disrupt both this diversity and this balance, giving rise to two characteristics: \textbf{1) complete attention hijacking}, where a prominent, patch trigger drives nearly all heads to attend to the trigger region, yielding abnormally high inter-head spatial overlap; and \textbf{2) partial attention hijacking}, in which a stealthier blended trigger affects heads unevenly, leading to abnormally high variance in attention concentration across heads. Together, these two characteristics cover the two dominant trigger families (patch-based and blended-based), providing a principled signal for detecting backdoor without requiring training data or model outputs.

Motivated by these findings, we propose \textbf{DetectViT}, a test-time backdoor detection method by examining the attention matrix of ViTs. For each test sample, DetectViT extracts the CLS-to-patch attention distributions across all heads and layers, and computes two complementary scores: an inter-head consistency score $\mathcal{S}_{\text{cons}}$ that captures \textbf{complete attention hijacking}, and an inter-head entropy variance score $\mathcal{S}_{\text{var}}$ that captures \textbf{partial attention hijacking}. Detection thresholds are estimated using a small out-of-distribution (OOD) calibration set drawn from any readily available dataset. A test sample is flagged as backdoored if either score exceeds its corresponding threshold.

Extensive experiments on four representative attacks across diverse ViT-based architectures (DeiT, CLIP, LLaVA) and downstream tasks (image classification and image captioning) demonstrate that DetectViT substantially outperforms existing baselines. For example, on BadCLIP, DetectViT achieves a true positive rate of 100\% at a false positive rate of only 0.17\% using as few as $128$ OOD calibration samples. Crucially, DetectViT introduces no additional inference latency since it uses attention weights already computed during inference, making it well-suited for real-world deployment. We summarize our main contributions below:

\begin{itemize}
\item We identify two anomalous attention phenomena induced by backdoors, i.e., \textbf{complete} and \textbf{partial attention hijacking}, characterized by abnormally high inter-head attention overlap and large inter-head variance in attention concentration. To our knowledge, this is the first systematic characterization of backdoor behavior through inter-head attention discrepancy.

\item We propose \textbf{DetectViT}, a test-time detection method by capturing these two phenomena via an inter-head consistency score and an inter-head entropy variance score. \textbf{DetectViT} requires neither training data, model outputs, nor any prior knowledge of the trigger.

\item Extensive experiments across diverse attacks, models, and tasks show that DetectViT achieves state-of-the-art performance with only a small OOD calibration set, remains robust to the source and size of the calibration data, and is highly effective against adaptive attacks, with negligible computation overhead.

\end{itemize}

\section{Related Work}
\textbf{Vision Transformers.}
Vision Transformers~\cite{dosovitskiy2020image} introduced the transformer architecture~\cite{vaswani2017attention} to visual representation learning by treating an image as a sequence of fixed-size patches and processing them with multi-head self-attention. Subsequent variants have demonstrated strong performance across diverse vision tasks~\cite{liu2021swin, wang2021pyramid,li2022more,pmlr-v139-touvron21a, li2023diverse, mehta2021mobilevit}. Among them, DeiT~\cite{pmlr-v139-touvron21a} extended ViTs by introducing a knowledge distillation strategy that enables efficient training without massive datasets. Building on them, CLIP~\cite{radford2021learning} proposed a vision-language model trained on large-scale image-text pairs via contrastive learning, inspiring a series of VLMs~\cite{li2022blip, yu2022coca}. LVLMs like LLaVA~\cite{liu2024improved} further advanced this direction by connecting a pretrained ViT-based vision encoder to a large language model, enabling image captioning and visual question answering, with numerous follow-up works adopting similar architectures~\cite{zhu2023minigpt, wang2024qwen2, li2024core, yang2025qwen3, team2025gemma}. 

\textbf{Backdoor Attacks on Vision Transformers.}
Backdoor attacks~\cite{gu2017badnets, chen2017targeted} can manipulate the victim model to produce attacker-specified predictions on trigger-bearing inputs while behaving normally on clean data. Recent works \cite{yuan2023you,zheng2023trojvit} have demonstrated that the self-attention mechanism of ViTs introduces vulnerabilities that can be exploited to craft more effective and stealthy backdoor attacks.
BadViT~\cite{yuan2023you} designs an optimized patch-based trigger that manipulates the attention to concentrate on the trigger region, achieving high attack success rates. TrojViT~\cite{zheng2023trojvit} optimizes a multiple-patch trigger on adaptive regions and takes a different approach by injecting the backdoor into the classification head rather than the encoder, making it harder to detect via encoder-level analysis. AGEB~\cite{guo2024backdoor} proposes an attention gradient-based erosion strategy that selectively corrupts pixels in regions of maximal attention gradient. AIBA~\cite{wang2025attention} follows TrojViT, optimizing a multiple-patch adversarial perturbation as the trigger.

Beyond classification, backdoor attacks have been extended to ViT-based VLMs. BadCLIP~\cite{liang2024badclip} targets the CLIP architecture by designing a trigger that manipulates both the visual and textual representations. More recently, BadVision~\cite{liu2025stealthy} reveals a new and more far-reaching threat: by compromising only the pretrained vision encoder, an adversary can induce significant visual hallucinations in downstream LVLMs such as LLaVA~\cite{liu2024improved} and MiniGPT-4~\cite{zhu2023minigpt} without touching the language model. Since pretrained vision encoders are widely shared and reused across many LVLMs, a single poisoned encoder can propagate backdoor behaviors to numerous downstream models. This highlights the urgent need for effective detection approaches that operate at the vision encoder level.

\textbf{Test-time Backdoor Samples Detection.}
Multiple methods detect backdoors requiring training data~\cite{feng2023detecting,pan2023asset,rong2025backdoor,yuanactivation}, making them inapplicable to protect third-party pretrained models. 
Test-time detection methods inspect suspicious samples at inference time without requiring training data. The pioneering work STRIP~\cite{gao2019strip} adds perturbations to test inputs and flags samples whose labels remain abnormally stable. TeCo~\cite{liu2023detecting} analyzes the consistency of hard labels under a series of image degradation transformations. Scale-Up~\cite{guoscale} amplifies the test image by varying scaling factors and detects samples based on their label changes. PatchProc~\cite{doan2023defending} processes images at the patch level and identifies backdoor samples via label shifts under patch-wise manipulations.
However, these label-based methods are fundamentally restricted to classification tasks and cannot handle open-ended outputs like image captioning. Advanced embedding-based methods avoid this restriction by operating on vision encoder representations: DeDe~\cite{hou2025dede} detects anomalous encoder mappings via a trained decoder; BackdoorIDS~\cite{huang2026backdoorids} clusters the embedding sequence under progressive masking to detect backdoor behavior, and SubspaceDetection~\cite{songtest} identifies poisoned CLIP images via shallow text-subspace alignment. However, these embedding-based methods are blind to backdoors injected into the classification head. 
Moreover, most existing test-time methods \cite{liu2023detecting, guoscale, doan2023defending, huang2026backdoorids, songtest} rely on generating a large number of augmented variants per test sample, incurring substantial computational overhead at inference time.
 
In contrast, DetectViT addresses the above limitations by operating on the internal multi-head attention behavior of ViTs, without relying on model outputs, detecting different types of triggers, and generalizing across classification and multimodal tasks with negligible additional time overhead.

\section{Motivation}
\label{sec:motivation}
The multi-head attention mechanism in ViTs is fundamentally designed to capture diverse and complementary visual representations: each head independently projects token embeddings into a separate subspace, allowing the model to simultaneously analyze different aspects of a visual scene. 
As shown in Fig.~\ref{moti_fig:a}, for a clean image, this design naturally induces \textbf{attention diversity and balance}: as each head perceives the image through a distinct lens, different heads attend to spatially different regions~\cite{caron2021emerging}; meanwhile, since all heads share the same architectural capacity and are optimized under the same objective, their concentration of the attention distribution remains broadly comparable.

We argue that backdoor triggers disrupt this diversity and balance. While prior work~\cite{yuan2023you,zheng2023trojvit,rong2025backdoor} has shown that triggers hijack heads' attention, our analysis further reveals that this hijacking is far from consistent across heads; rather, it manifests in two distinct regimes depending on the perceptibility and spatial extent of the trigger, as empirically corroborated in Sec.~\ref{section:experiment_results} (Fig. \ref{fig:per_layer_signals}):

\begin{wrapfigure}{r}{0.62\linewidth}
\vspace{-17pt}
	\centering
	\subfloat[Clean image]{\label{moti_fig:a}\includegraphics[width=0.20\textwidth]{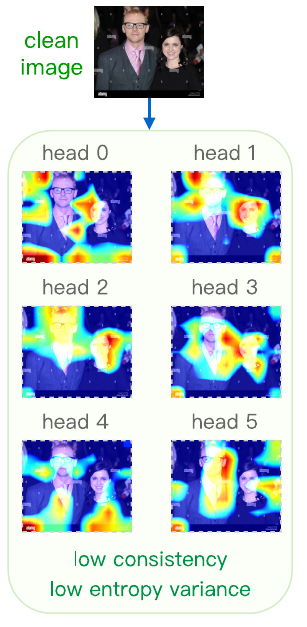}}\hspace{-1mm}
	\subfloat[Complete hijacking]{\label{moti_fig:b}\includegraphics[width=0.20\textwidth]{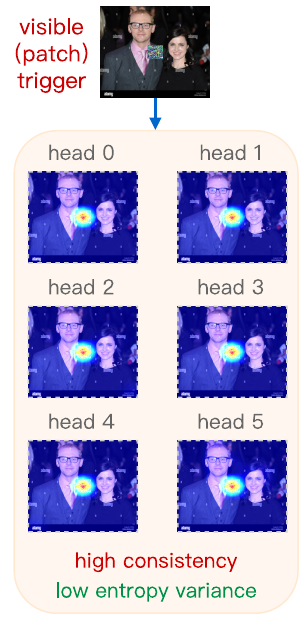}}\hspace{-1mm}
        \subfloat[Partial hijacking]{\label{moti_fig:c}\includegraphics[width=0.20\textwidth]{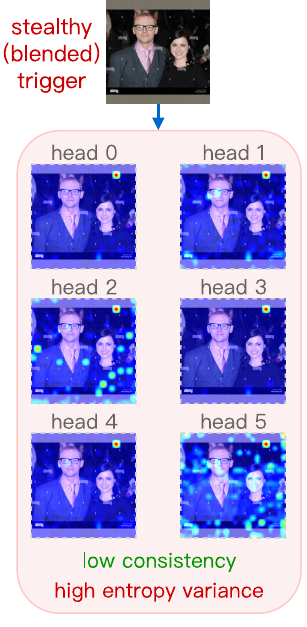}}
	\caption{Visualization of the head attention distribution over the CLS token of clean/backdoor images.}
        \label{visulization}
\vspace{-10pt}
\end{wrapfigure}
\textbf{1. Complete attention hijacking} occurs when a prominent, localized trigger is present. The trigger pattern introduces an anomalously dominant local feature that overrides the learned diversity of the multi-head mechanism, forcing all attention heads to converge onto the trigger region regardless of their respective subspace projections. As visualized in Fig.~\ref{moti_fig:b}, the attention maps of all heads collapse onto a single small region corresponding to the trigger, exhibiting an inter-head spatial overlap that stands in sharp contrast to the diverse distributions observed on clean images. 
    
\textbf{2. Partial attention hijacking} characterizes the behavior of stealthy, low-visibility triggers, such as blended-based attacks, where a global pattern is superimposed at low intensity across the entire image. The backdoor is activated by the cumulative contribution of many faint, image-wide patches rather than any single dominant region; consequently, different heads respond to the trigger in fundamentally different ways, as shown in Fig.~\ref{moti_fig:c}: some heads are fully hijacked and attend broadly across nearly all trigger-bearing patches (e.g., heads~2 and~5), while others are only partially captured and concentrate narrowly on the most salient trigger patches (e.g., heads~0 and~3). This heterogeneous response produces a pronounced disparity in the concentration of the attention distribution across each head, whereas clean images yield the comparable concentration.


Taken together, these two forms of attention hijacking provide a principled and model-agnostic foundation for detecting backdoor samples across both visible and stealthy trigger modalities, requiring neither access to training data nor any assumption about the specific trigger pattern employed.

\section{Method}
\begin{figure}[t]
    \centering
    \includegraphics[width=1\textwidth]{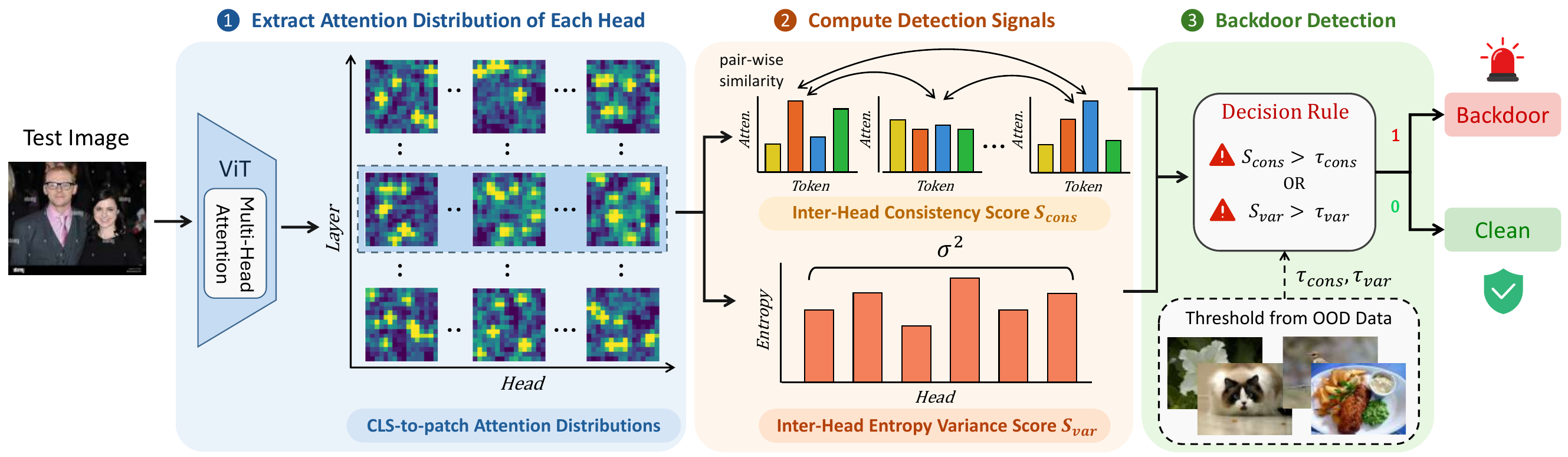}
    \caption{Overview of our proposed DetectViT.}
    \label{fig:pipeline}
\end{figure}

We present \textbf{DetectViT}, a novel test-time framework for detecting backdoor samples in ViTs. The method exploits two complementary deviations in multi-head attention behavior induced by backdoor triggers: abnormal inter-head spatial overlap under complete hijacking, and abnormal inter-head concentration variance under partial hijacking. 
The overall pipeline is illustrated in Fig.~\ref{fig:pipeline}.

\subsection{Attention Distribution of Each Head over the CLS Token}
Given an input image $x$, we pass it through the ViT encoder and extract the attention weights produced at each transformer layer. Formally, let $L$ denote the total number of layers and $H$ the number of attention heads per layer. 
At layer $\ell \in \{1, \dots, L\} $, let $\mathbf{A}^{(\ell,h)}  \in \mathbb{R}^{(N+1) \times (N+1)}$ denote the attention weight matrix for head $h$, where $N$ is the number of patch tokens, and the additional token corresponds to the CLS token.

We focus on the attention that the CLS token pays to each patch token, as it encodes the global semantic representation of the image. We extract the CLS-to-patch attention vector for each head:
\begin{equation} \label{eq:cls_attn}
    \mathbf{a}^{(\ell,h)} = \mathbf{A}^{(\ell,h)}_{0,1:N} \in \mathbb{R}^{N},
\end{equation}
where the subscript $0$ indexes the CLS token row and $1{:}N$ indexes the $N$ patch token columns. We then normalize each vector to form a proper probability distribution:
\begin{equation} \label{eq:normalize}
    \tilde{\mathbf{a}}^{(\ell,h)} = \frac{\mathbf{a}^{(\ell,h)}}{\sum_{i=1}^{N} a^{(\ell,h)}_i} \in \mathbb{R}^{N}.
\end{equation}
The resulting $\tilde{\mathbf{a}}^{(\ell,h)}$ represents the normalized attention distribution of head $h$ at layer $\ell$ over all patch tokens, and serves as the basis for the two detection signals described below.

\subsection{Detection Signals}

\subsubsection{Inter-Head Consistency Score}

To detect complete attention hijacking, wherein all heads converge onto the same trigger region, we quantify the resulting inter-head spatial overlap via the pairwise cosine similarity between the attention distributions of different heads within the same layer. For any pair of heads $(h, h')$ at layer $\ell$, the cosine similarity is:
\begin{equation} \label{eq:cosine}
    \mathrm{sim}^{(\ell)}(h, h') = \frac{\tilde{\mathbf{a}}^{(\ell,h)} \cdot \tilde{\mathbf{a}}^{(\ell,h')}}{\|\tilde{\mathbf{a}}^{(\ell,h)}\|_2 \cdot \|\tilde{\mathbf{a}}^{(\ell,h')}\|_2}.
\end{equation}
We then aggregate these into a layer-wise 
inter-head consistency score as the mean over all pairs:
\begin{equation} \label{eq:layer_cons}
    c^{(\ell)} = \frac{1}{\binom{H}{2}} \sum_{h < h'} \mathrm{sim}^{(\ell)}(h, h').
\end{equation}
A high value of $c^{(\ell)}$ indicates that all heads are attending to the same spatial region, which is the hallmark of complete attention hijacking. The final consistency score for the sample is taken as the maximum across all layers:
\begin{equation} \label{eq:score_cons}
    \mathcal{S}_{\text{cons}} = \max_{\ell} \; c^{(\ell)}.
\end{equation}

\subsubsection{Inter-Head Entropy Variance Score}

To detect partial attention hijacking, wherein different heads exhibit divergent degrees of attention concentration, we quantify the concentration of each head's attention distribution by Shannon entropy:
\begin{equation} \label{eq:entropy}
    \mathcal{H}^{(\ell,h)} = -\sum_{i=1}^{N} \tilde{a}^{(\ell,h)}_i \log\!\left(\tilde{a}^{(\ell,h)}_i\right).
\end{equation}
A low entropy value indicates that the head concentrates its attention on a small number of patches (i.e., highly focused), while a high entropy value indicates diffuse attention over many patches. Under partial hijacking, different heads respond to the trigger with heterogeneous concentration, producing an abnormally high variance in entropy across heads within a layer, which we refer to as the inter-head entropy variance score:
\begin{equation} \label{eq:layer_var}
    v^{(\ell)} = \mathrm{Var}_{h}\!\left(\mathcal{H}^{(\ell,h)}\right) = \frac{1}{H} \sum_{h=1}^{H} \left(\mathcal{H}^{(\ell,h)} - \bar{\mathcal{H}}^{(\ell)}\right)^2,
\end{equation}
where $\bar{\mathcal{H}}^{(\ell)} = \frac{1}{H}\sum_{h}\mathcal{H}^{(\ell,h)}$ is the mean entropy at layer $\ell$. The final entropy variance score is:
\begin{equation} \label{eq:score_var}
    \mathcal{S}_{\text{var}} = \max_{\ell} \, v^{(\ell)}.
\end{equation}

\subsection{Threshold Estimation via OOD Calibration}

Instead of relying on in-distribution training data, we estimate detection thresholds from a small set of OOD samples that are presumed to be clean. In practice, we find that as few as $M = 64$ or $128$ such samples suffice. For each calibration sample $x_m$, we compute both scores $\mathcal{S}_{\text{cons}}(x_m)$ and $\mathcal{S}_{\text{var}}(x_m)$. We then estimate the empirical mean and standard deviation of each score over the calibration set:
\begin{align}
    \mu_{\text{cons}} &= \frac{1}{M}\sum_{m=1}^{M} \mathcal{S}_{\text{cons}}(x_m), \quad
    \sigma_{\text{cons}} = \mathrm{Std}_{m}\!\left(\mathcal{S}_{\text{cons}}(x_m)\right), \label{eq:stats_cons} \\
    \mu_{\text{var}}  &= \frac{1}{M}\sum_{m=1}^{M} \mathcal{S}_{\text{var}}(x_m),  \quad
    \sigma_{\text{var}}  = \mathrm{Std}_{m}\!\left(\mathcal{S}_{\text{var}}(x_m)\right). \label{eq:stats_var}
\end{align}
Since the scores of clean samples approximately follow a Gaussian distribution (verified in Appendix~\ref{app:gaussian}), we apply the three-sigma rule to set the detection thresholds:
\begin{equation} \label{eq:threshold}
    \tau_{\text{cons}} = \mu_{\text{cons}} + 3\sigma_{\text{cons}}, \qquad
    \tau_{\text{var}}  = \mu_{\text{var}}  + 3\sigma_{\text{var}}.
\end{equation}
This ensures that the probability of a clean sample being falsely flagged as backdoor is less than $0.15\%$ under the one-sided three-sigma rule, providing a conservative false positive rate.

Importantly, our method does not restrict the source of the OOD calibration samples. We empirically verify that thresholds estimated from a diverse range of OOD datasets yield consistently strong detection performance, including iNaturalist~\cite{inaturalist-2021}, Food101~\cite{bossard14}, Oxford-IIIT Pet~\cite{parkhi12a}, Places365~\cite{zhou2017places}, STL-10~\cite{coates2011analysis}, Flowers102~\cite{Nilsback08}, and PASCAL VOC~\cite{Everingham10}. This flexibility makes the method practical in real-world scenarios where access to in-distribution clean data may be restricted.

\subsection{Test-time Backdoor Sample Detection}

Given a test sample $x$, we compute both detection scores and compare them against the estimated thresholds. The sample is classified as backdoored if either score exceeds its corresponding threshold:
\begin{equation} \label{eq:decision}
    \hat{y} =
    \begin{cases}
        1 \; (\text{backdoor}) & \text{if } \mathcal{S}_{\text{cons}}(x) > \tau_{\text{cons}}
                                  \;\text{ or }\; \mathcal{S}_{\text{var}}(x) > \tau_{\text{var}}, \\
        0 \; (\text{clean})    & \text{otherwise.}
    \end{cases}
\end{equation}
The use of a disjunctive criterion ensures that both trigger modalities, complete hijacking (captured by $\mathcal{S}_{\text{cons}}$) and partial hijacking (captured by $\mathcal{S}_{\text{var}}$), are independently detectable, making the method robust to diverse attack strategies. The algorithm for DetectViT is presented in Appendix \ref{appendix:algorithm}.

\section{Experimental Setup}

\paragraph{Attacked Models and Datasets.}
We evaluate DetectViT against four representative backdoor attacks spanning diverse ViT architectures and downstream tasks. Specifically, we consider BadViT~\cite{yuan2023you} and TrojViT~\cite{zheng2023trojvit}, both of which employ optimized patch-based triggers, targeting image classification models based on 
DeiT-T and DeiT-S~\cite{pmlr-v139-touvron21a}, evaluated on 
Caltech-101~\cite{li_andreeto_ranzato_perona_2022} and Tiny ImageNet~\cite{le2015tiny}, respectively. We further consider BadCLIP~\cite{liang2024badclip}, which also employs an optimized patch-based trigger to attack the CLIP~\cite{radford2021learning} vision-language model based ViT-B-32 \cite{dosovitskiy2020image}, evaluated on ImageNet-1K~\cite{deng2009imagenet}, and BadVision~\cite{liu2025stealthy}, which employs a blended-based trigger to poison the vision encoder of LLaVA-1.5~\cite{liu2024improved} (using clip-vit-large-patch14-336), evaluated on COCO Caption~\cite{chen2015microsoft}. 
We poison $5\%$ of the test data by default. More details can be seen in Appendix \ref{appendix:attack_details}.

\paragraph{Baselines.}
We compare DetectViT against five competitive backdoor sample detection methods. TeCo~\cite{liu2023detecting}, Scale-Up~\cite{guoscale}, and PatchProc~\cite{doan2023defending} operate on the model's output label space, flagging backdoor samples by analyzing prediction consistency or sensitivity under input transformations. DeDe~\cite{hou2025dede}, and BackdoorIDS~\cite{huang2026backdoorids} operate in the model's embedding space, detecting anomalous feature representations induced by the trigger. Our method is orthogonal to both paradigms, operating on the attention weights without requiring either label-level queries or embedding-level access.

\paragraph{Evaluation Metrics.}
We evaluate all methods using three standard binary classification metrics: True Positive Rate (TPR), False Positive Rate (FPR), and the Area Under the ROC Curve (AUC). TPR measures the fraction of backdoor samples correctly identified, FPR measures the fraction of clean samples incorrectly flagged, and AUC provides a threshold-independent summary of overall detection performance. Higher TPR and AUC, and lower FPR, indicate better detection.

\paragraph{Hyperparameters.}
Our method introduces minimal hyperparameters. The only configurable parameters are the source and size of the OOD calibration set used for threshold estimation. Unless otherwise specified, we use iNaturalist~\cite{inaturalist-2021} as the default OOD source with $M = 128$ calibration samples. We further conduct ablation studies on both the choice of OOD dataset and the value of $M$, demonstrating that our method is robust to both choices across the seven OOD datasets. All experiments are conducted on a single NVIDIA GeForce RTX 4090 GPU. 
 

\section{Experimental Results}
\label{section:experiment_results}
\begin{table}[t]
\centering
\caption{Comparison of backdoor sample detection methods. TPR and FPR are reported as percentages (\%). AUC is reported on a $[0,1]$ scale. We bold the best value.}
\label{tab:main}
\resizebox{\textwidth}{!}{%
\begin{tabular}{@{}lcccccccccccc@{}}
\toprule
\multicolumn{1}{l}{\multirow{2}{*}{Defense}} & \multicolumn{3}{c}{BadViT}                      & \multicolumn{3}{c}{TrojViT}                    & \multicolumn{3}{c}{BadCLIP}                     & \multicolumn{3}{c}{BadVision}                  \\ \cmidrule(lr){2-4} \cmidrule(lr){5-7} \cmidrule(lr){8-10} \cmidrule(lr){11-13} 
\multicolumn{1}{c}{}                         & TPR $\uparrow$             & FPR $\downarrow$           & AUC $\uparrow$           & TPR $\uparrow$            & FPR $\downarrow$           & AUC $\uparrow$           & TPR $\uparrow$             & FPR $\downarrow$           & AUC $\uparrow$           & TPR $\uparrow$            & FPR $\downarrow$           & AUC $\uparrow$           \\ \midrule
\rowcolor{gray!10}TeCo~\cite{liu2023detecting}                                         & 77.95           & 61.86         & 0.63          & 55.00          & 51.76         & 0.51          & 99.20           & 39.52         & 0.89          & -              & -             & -             \\
Scale-up~\cite{guoscale}                                     & 88.65           & 30.66         & 0.81          & 85.00          & 41.98         & 0.74          & 91.84           & 65.03         & 0.66          & -              & -             & -             \\
\rowcolor{gray!10}PatchProc~\cite{doan2023defending}                                    & 90.61           & 18.42         & 0.89          & 87.60          & 21.44         & 0.85          & 86.97           & 26.78         & 0.82          & -              & -             & -             \\
DeDe~\cite{hou2025dede}                                         & 63.97           & 23.29         & 0.78          & 50.20          & 25.80         & 0.72          & 55.12           & 27.97         & 0.69          & 7.00           & 29.68         & 0.50          \\
\rowcolor{gray!10}B-IDS~\cite{huang2026backdoorids}                                  & 80.79           & 9.73          & -             & 5.00           & 10.84         & -             & 93.24           & 16.88         & -             & \textbf{96.00} & 15.96         & -             \\
\textbf{DetectViT}                              & \textbf{100.00} & \textbf{1.58} & \textbf{1.00} & \textbf{90.80} & \textbf{5.94} & \textbf{0.95} & \textbf{100.00} & \textbf{0.17} & \textbf{1.00} & \textbf{96.00}          & \textbf{3.00} & \textbf{0.99} \\ \bottomrule
\end{tabular}
}
\end{table}

\paragraph{Detection Performance Compared to Baselines.}
Table~\ref{tab:main} presents the detection performance of all methods across four backdoor attacks. DetectViT achieves the best results across all four attacks. 
BackdoorIDS assigns discrete cluster labels rather than continuous scores and thus does not yield an AUC. Further, the limitations of existing baselines are systematic rather than incidental. Label-space methods (TeCo, Scale-Up, PatchProc) are fundamentally constrained to classification tasks and are entirely inapplicable to BadVision, since LLaVA-1.5 produces free-form text output rather than discrete class labels. Embedding-space methods (DeDe, BackdoorIDS) avoid this restriction but fail on TrojViT (TPRs of 50.20\% and 5.00\%), since its backdoor is injected into the classification head rather than the encoder, leaving poisoned embeddings indistinguishable from clean ones. DetectViT sidesteps both failure modes by operating directly on attention weights rather than model outputs or embeddings, enabling robust detection across all settings.
Beyond the attacks designed specifically for vision transformers, we further evaluate five additional attacks covering self-supervised encoders (BadEncoder~\cite{jia2022badencoder}, DRUPE~\cite{tao2024distribution}), clean-label poisoning (Label-Consistent~\cite{turner2019label}), frequency-domain triggers (Low-Frequency~\cite{zeng2021rethinking}), and optimized global triggers (Blended-Optimized~\cite{chen2017targeted}), where DetectViT is the only method that maintains an AUC of at least 0.97 across all of them (Appendix~\ref{app:more_attacks}).


\begin{table}[t]
\centering
\caption{Detection performance under different OOD calibration sources and
sizes.}
\label{tab:ood_source}
\resizebox{0.92\textwidth}{!}{%
\begin{tabular}{@{}lccccccccc@{}}
\toprule
\multirow{3}{*}{OOD Source} & \multirow{3}{*}{$M$}
    & \multicolumn{2}{c}{BadViT}
    & \multicolumn{2}{c}{TrojViT}
    & \multicolumn{2}{c}{BadCLIP}
    & \multicolumn{2}{c}{BadVision} \\
\cmidrule(lr){3-4} \cmidrule(lr){5-6} \cmidrule(lr){7-8} \cmidrule(lr){9-10}
& & TPR $\uparrow$ & FPR $\downarrow$ & TPR $\uparrow$ & FPR $\downarrow$ & TPR $\uparrow$ & FPR $\downarrow$ & TPR $\uparrow$ & FPR $\downarrow$ \\
\midrule
\multirow{2}{*}{iNaturalist~\cite{inaturalist-2021}}
    & \cellcolor{gray!10}64  & \cellcolor{gray!10}100.00 & \cellcolor{gray!10}2.08           & \cellcolor{gray!10}93.40 & \cellcolor{gray!10}13.82          & \cellcolor{gray!10}100.00 & \cellcolor{gray!10}1.12           & \cellcolor{gray!10}96.00 & \cellcolor{gray!10}2.42          \\
    & 128 & 100.00 & 1.58  & 90.80 & 5.94  & 100.00 & 0.17  & 96.00 & 3.00          \\
\midrule
\multirow{2}{*}{Food101~\cite{bossard14}}
    & \cellcolor{gray!10}64  & \cellcolor{gray!10}100.00 & \cellcolor{gray!10}10.37          & \cellcolor{gray!10}94.60 & \cellcolor{gray!10}20.33          & \cellcolor{gray!10}100.00 & \cellcolor{gray!10}7.52           & \cellcolor{gray!10}96.00 & \cellcolor{gray!10}3.42          \\
    & 128 & 100.00 & 9.39           & 92.60 & 11.81          & 100.00 & 8.52           & 96.00 & 2.95          \\
\midrule
\multirow{2}{*}{Oxford-IIIT Pet~\cite{parkhi12a}}
    & \cellcolor{gray!10}64  & \cellcolor{gray!10}100.00 & \cellcolor{gray!10}5.48           & \cellcolor{gray!10}88.60 & \cellcolor{gray!10}9.04           & \cellcolor{gray!10}100.00 & \cellcolor{gray!10}8.90           & \cellcolor{gray!10}97.00 & \cellcolor{gray!10}5.16          \\
    & 128 & 100.00 & 2.94           & 91.20 & 5.78           & 100.00 & 6.91           & 96.00 & 3.26          \\
\midrule
\multirow{2}{*}{Places365~\cite{zhou2017places}}
    & \cellcolor{gray!10}64  & \cellcolor{gray!10}100.00 & \cellcolor{gray!10}5.30           & \cellcolor{gray!10}91.60 & \cellcolor{gray!10}8.60           & \cellcolor{gray!10}100.00 & \cellcolor{gray!10}5.47           & \cellcolor{gray!10}92.00 & \cellcolor{gray!10}0.47          \\
    & 128 & 100.00 & 3.75           & 91.20 & 6.47           & 100.00 & 6.34           & 92.00 & 0.32 \\
\midrule
\multirow{2}{*}{STL-10~\cite{coates2011analysis}}
    & \cellcolor{gray!10}64  & \cellcolor{gray!10}100.00 & \cellcolor{gray!10}3.82           & \cellcolor{gray!10}89.20 & \cellcolor{gray!10}10.39          & \cellcolor{gray!10}100.00 & \cellcolor{gray!10}9.20           & \cellcolor{gray!10}97.00 & \cellcolor{gray!10} 5.84          \\
    & 128 & 100.00 & 2.92           & 89.20 & 9.78           & 100.00 & 6.92           & 97.00 & 5.16          \\
\midrule
\multirow{2}{*}{Flowers102~\cite{Nilsback08}}
    & \cellcolor{gray!10}64  & \cellcolor{gray!10}100.00 & \cellcolor{gray!10}5.39           & \cellcolor{gray!10}92.00 & \cellcolor{gray!10}7.61           & \cellcolor{gray!10}100.00 & \cellcolor{gray!10}7.11           & \cellcolor{gray!10}98.00 & \cellcolor{gray!10}9.15         \\
    & 128 & 100.00 & 3.80           & 89.20 & 3.66           & 100.00 & 5.63           & 97.00 & 5.84         \\
\midrule
\multirow{2}{*}{PASCAL VOC~\cite{Everingham10}}
    & \cellcolor{gray!10}64  & \cellcolor{gray!10}100.00 & \cellcolor{gray!10}2.61  & \cellcolor{gray!10}94.20 & \cellcolor{gray!10}18.21          & \cellcolor{gray!10}100.00 & \cellcolor{gray!10}4.16           & \cellcolor{gray!10}91.00 & \cellcolor{gray!10}0.16 \\
    & 128 & 100.00 & 1.92           & 93.00 & 13.73          & 100.00 & 2.65           & 91.00 & 0.16 \\
\bottomrule
\end{tabular}%
}
\end{table}

\begin{figure}[t]
    \centering
    \begin{minipage}{0.49\textwidth}
        \centering
        \includegraphics[width=\linewidth]{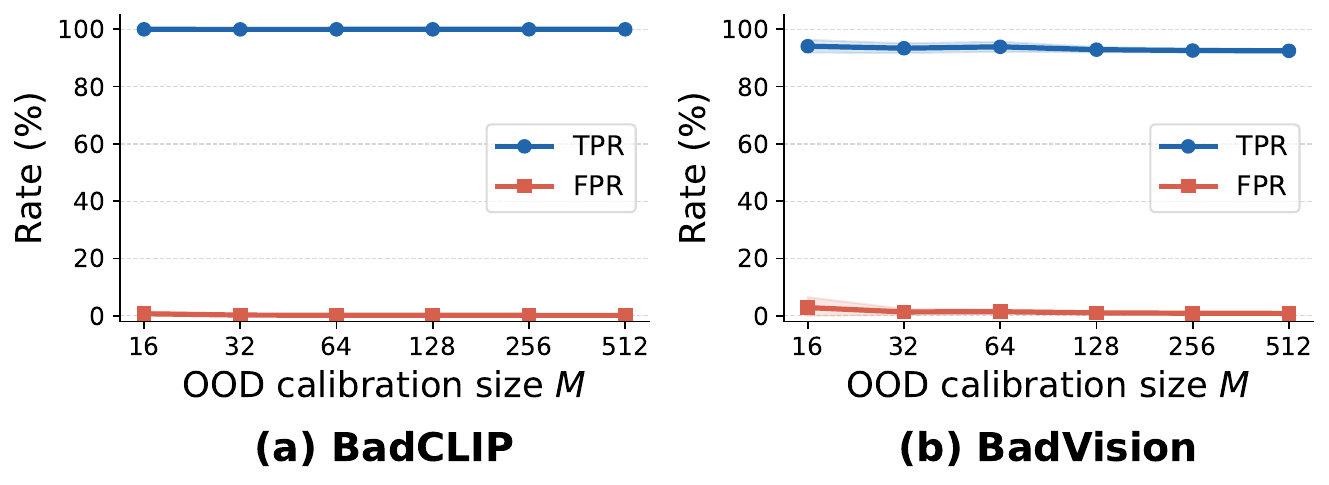} 
        \caption{Effectiveness of DetectViT under different numbers of OOD calibration samples $M$.}
        \label{fig:ood_size}
    \end{minipage}
    \begin{minipage}{0.49\textwidth}
        \centering
        \includegraphics[width=\linewidth]{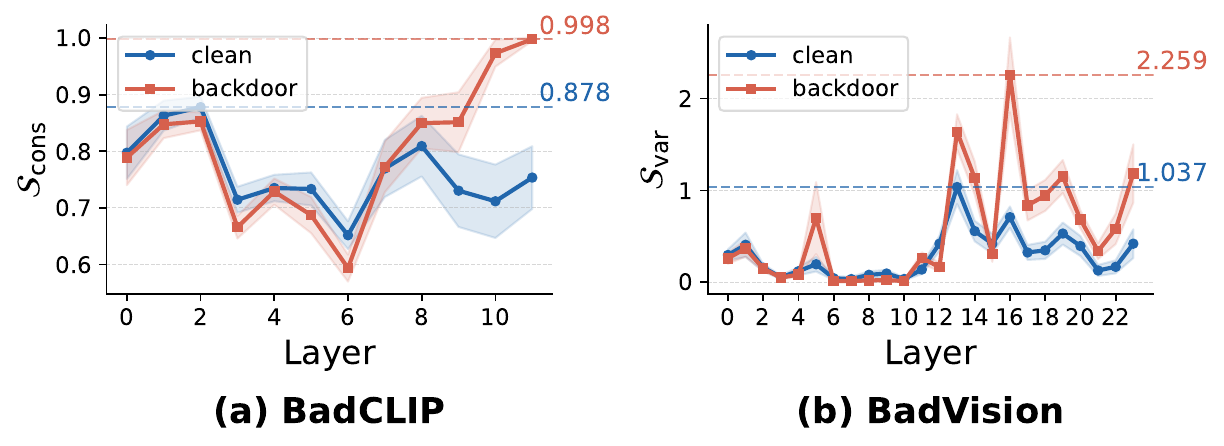} 
        \caption{Visualization of detection signals across all backdoor/clean samples.}
        \label{fig:per_layer_signals}
    \end{minipage}
\end{figure}

\paragraph{Detection Performance under Different OOD Calibration Data Sources.}
 Table~\ref{tab:ood_source} reports detection performance under seven different OOD calibration sources at $M = 64$ and $M = 128$. As the calibration data only influences the detection threshold, while not influencing the AUC value, we do not report it here.  
 Across all datasets and both sizes, TPR remains consistently high — DetectViT achieves $100\%$ TPR on BadViT and BadCLIP regardless of the OOD source, and maintains strong TPR on TrojViT and BadVision across all settings. FPR shows somewhat greater variation across sources, which is expected given that different OOD datasets have different distributional properties that affect threshold estimation; nevertheless, the FPR remains acceptably low for the vast majority of configurations. Notably, increasing $M$ from $64$ to $128$ consistently reduces FPR across nearly all source-attack combinations, echoing the trend observed in Fig.~\ref{fig:ood_size}. Despite the performance variation across different OOD sources, DetectViT consistently and substantially outperforms all baselines reported in Table~\ref{tab:main}. Overall, these results demonstrate that DetectViT is robust to the choice of OOD calibration source, and that practitioners can freely select from a wide range of off-the-shelf datasets without sacrificing detection reliability.

\paragraph{Detection Performance varying OOD Calibration Set Size.}
Fig.~\ref{fig:ood_size} illustrates how detection performance varies with the number of OOD calibration samples $M$. Each experiment is repeated under $10$ different random seeds, where the lines represent the mean value, and the shaded regions indicate the standard deviation.
We find that even with as few as $M = 16$ samples, DetectViT already achieves strong detection performance, demonstrating that the proposed threshold estimation is effective even in extremely data-scarce scenarios. Further, the variance of both TPR and FPR decreases rapidly as $M$ increases: at $M = 16$, the shaded regions remain noticeable, whereas at $M = 32$ or $64$, the variance has largely converged to a negligible level. This convergence confirms that our method requires only a small number of OOD samples to produce reliable and stable thresholds, making it practical in real-world scenarios where clean calibration data is scarce. Complete results for all attacks are provided in Appendix \ref{appendix:ood_size}.

\paragraph{Visualization of Detection Signals.} Fig.~\ref{fig:per_layer_signals} visualizes the per-layer detection signals averaged over all clean and backdoor samples, where the lines and shaded regions represent the mean value and the standard deviation, respectively. The peak signal value of backdoor samples substantially exceeds that of clean samples, whether measured by inter-head consistency $\mathcal{S}_{\text{cons}}$ on BadCLIP or entropy variance $\mathcal{S}_{\text{var}}$ on BadVision, confirming that the two proposed signals provide a reliable and discriminative separation between backdoor and clean inputs. Moreover, the peak signal tends to occur in the deeper layers of the model, suggesting that backdoor features are more prominently represented in posterior transformer layers. Complete results are provided in Appendix \ref{appendix:visualization_signals}.

\begin{wraptable}{r}{0.55\linewidth}
\vspace{-12pt}
\centering
\caption{Average per-sample detection time (s)  on each attacked
model. We bold the best value.}
\label{tab:efficiency}
\resizebox{0.55\columnwidth}{!}{
\begin{tabular}{@{}lcccc@{}}
\toprule
Defense & BadViT & TrojViT & BadCLIP & BadVision \\
\midrule
\rowcolor{gray!10}TeCo~\cite{liu2023detecting}               & 2.5094          & 4.2920          & 1.8448          & -               \\
Scale-Up~\cite{guoscale}        & 0.0246          & 0.0261          & 0.0696          & -               \\
\rowcolor{gray!10}PatchProc~\cite{doan2023defending}     & 0.6686          & 0.6926          & 0.8257          & -               \\
DeDe~\cite{hou2025dede}               & 0.0206          & 0.0279          & 0.1034          & 0.6060          \\
\rowcolor{gray!10}B-IDS~\cite{huang2026backdoorids} & 0.1734          & 0.1982          & 0.2560          & 3.3458          \\
\textbf{DetectViT}               & \textbf{0.0124} & \textbf{0.0134} & \textbf{0.0163} & \textbf{0.0226} \\
\bottomrule
\end{tabular}
}
\end{wraptable}
\paragraph{Runtime of Detection Methods.}
Table~\ref{tab:efficiency} reports the average per-sample detection time (in seconds) of all methods evaluated over $100$ test samples on each attacked model. DetectViT achieves the lowest detection time across all four settings. Most baselines incur substantially higher latency because they rely on augmenting each test sample into multiple transformed copies before analysis. In particular, TeCo additionally applies computationally expensive image perturbations such as blurring, resulting in detection times of up to $4.29$s per sample, more than $300\times$ slower than DetectViT on TrojViT. DeDe achieves competitive per-sample inference time owing to its lightweight decoder, but this comes at the cost of a significant one-time training overhead to train the decoder. Notably, the reported runtime \emph{upper-bounds} DetectViT's true detection cost: the bulk is the forward pass, while computing cosine similarities and entropies over the resulting attention weights is negligible by comparison. In practice, this forward pass overhead overlaps almost entirely with standard inference.

\begin{wraptable}{r}{0.72\linewidth}
\vspace{-12pt}
\centering
\caption{Ablation study on the two detection signals. \checkmark indicates the corresponding signal is enabled. We bold the best value.}
\label{tab:ablation}
\resizebox{0.72\columnwidth}{!}{%
\begin{tabular}{@{}cccccccccc@{}}
\toprule
\multicolumn{2}{c}{Signal} & \multicolumn{2}{c}{BadViT} & \multicolumn{2}{c}{TrojViT} & \multicolumn{2}{c}{BadCLIP} & \multicolumn{2}{c}{BadVision} \\
\cmidrule(lr){1-2} \cmidrule(lr){3-4} \cmidrule(lr){5-6} \cmidrule(lr){7-8} \cmidrule(lr){9-10}
$\mathcal{S}_{\text{cons}}$ & $\mathcal{S}_{\text{var}}$
    & TPR $\uparrow$ & FPR $\downarrow$ & TPR $\uparrow$ & FPR $\downarrow$ & TPR $\uparrow$ & FPR $\downarrow$ & TPR $\uparrow$ & FPR $\downarrow$ \\
\midrule
\rowcolor{gray!10}\checkmark &            & \textbf{100.00} & \textbf{0.54}          & 90.40 & 4.22          & \textbf{100.00} & \textbf{0.02}          & 0.00  & \textbf{0.00}          \\
           & \checkmark & 80.35  & 1.06          & 44.20 & \textbf{1.76}          & 89.52  & 0.16          & \textbf{96.00} & 3.00          \\
\rowcolor{gray!10}\checkmark & \checkmark & \textbf{100.00} & 1.58 & \textbf{90.80} & 5.94 & \textbf{100.00} & 0.17 & \textbf{96.00} & 3.00 \\
\bottomrule
\end{tabular}%
}
\end{wraptable}
\paragraph{Ablation Study.}

Table~\ref{tab:ablation} reports the contribution of each detection signal of DetectViT.
For the three patch-based attacks (BadViT, TrojViT, BadCLIP), the consistency signal alone achieves near-perfect detection, confirming that complete attention hijacking, wherein all heads converge onto the trigger region, produces a strong and reliable inter-head consistency anomaly. Interestingly, for BadViT and BadCLIP, penalizing $\mathcal{S}_{\text{cons}}$ alone inadvertently induces high inter-head entropy variance, so $\mathcal{S}_{\text{var}}$ takes over, and DetectViT still achieves full detection while the attack remains near-perfect. In contrast, for the blended-based attack BadVision, the consistency signal completely fails, as the stealthy trigger is insufficiently salient to force all heads to converge onto the same region. The entropy variance signal, however, achieves $96.00\%$ TPR with a $3.00\%$ FPR, demonstrating that partial attention hijacking is exclusively captured by the concentration disparity across heads. 
The results confirm that the two signals are complementary and their combination is effective across both trigger modalities.


\begin{wraptable}{r}{0.5\linewidth}
\vspace{-12pt}
\centering
\caption{Performance under adaptive attacks.``w/o'' denotes training
without adaptive regularization.}
\label{tab:adaptive}
\resizebox{0.5\columnwidth}{!}{%
\begin{tabular}{@{}lccccc@{}}
\toprule
Attack & Adaptive & TPR $\uparrow$ & FPR $\downarrow$ & ASR $\uparrow$ & CA $\uparrow$ \\
\midrule
\multirow{3}{*}{BadViT}
    & \cellcolor{gray!10}w/o              & \cellcolor{gray!10}100.00 & \cellcolor{gray!10}1.58 & \cellcolor{gray!10}100.00 & \cellcolor{gray!10}83.04 \\
    & $\mathcal{S}_{\text{cons}}$ & 100.00 & 1.88 & 99.84  & 81.67 \\
    & \cellcolor{gray!10} $\mathcal{S}_{\text{cons}}$ + $\mathcal{S}_{\text{var}}$  & \cellcolor{gray!10}100.00 & \cellcolor{gray!10}1.32 & \cellcolor{gray!10}100.00 & \cellcolor{gray!10}40.66 \\
\midrule
\multirow{3}{*}{TrojViT}
    & w/o              & 90.80 & 5.94 & 100.00 & 85.17 \\
    & \cellcolor{gray!10} $\mathcal{S}_{\text{cons}}$ & \cellcolor{gray!10} 90.80 & \cellcolor{gray!10}5.94 & \cellcolor{gray!10}100.00 & \cellcolor{gray!10}85.17 \\
    & $\mathcal{S}_{\text{cons}}$ + $\mathcal{S}_{\text{var}}$  & 90.80 & 5.94 & 100.00 & 85.17 \\
\midrule
\multirow{3}{*}{BadCLIP}
    & \cellcolor{gray!10}w/o              & \cellcolor{gray!10}100.00 & \cellcolor{gray!10}0.17 & \cellcolor{gray!10}99.99 & \cellcolor{gray!10}61.39 \\
    & $\mathcal{S}_{\text{cons}}$ & 100.00 & 1.95 & 99.96  & 61.06 \\
    & \cellcolor{gray!10} $\mathcal{S}_{\text{cons}}$ + $\mathcal{S}_{\text{var}}$  & \cellcolor{gray!10}0.32   & \cellcolor{gray!10}0.61 & \cellcolor{gray!10}14.93  & \cellcolor{gray!10}61.42 \\
\midrule
\multirow{3}{*}{BadVision}
    & w/o   & 96.00 & 3.00 & 99.00 & 85.40 \\
    & \cellcolor{gray!10} $\mathcal{S}_{\text{var}}$ & \cellcolor{gray!10}0.00  & \cellcolor{gray!10}0.00 & \cellcolor{gray!10}35.00  & \cellcolor{gray!10}85.20 \\
    & $\mathcal{S}_{\text{cons}}$ + $\mathcal{S}_{\text{var}}$  & 0.00  & 0.26 & 11.00  & 83.00 \\
\bottomrule
\end{tabular}%
}
\end{wraptable}
\paragraph{Adaptive Attacks.}
We evaluate DetectViT against an adaptive adversary who penalizes our detection signals during backdoor training: $\mathcal{S}_{\text{cons}}$ for patch-based attacks and $\mathcal{S}_{\text{var}}$ for blended-based attacks, or both simultaneously. Table~\ref{tab:adaptive} reports TPR, FPR, Attack Success Rate (ASR), and Clean Accuracy (CA), where the CA of BadVision represents the CIDEr score \cite{vedantam2015cider}.
For BadViT and BadCLIP, penalizing $\mathcal{S}_{\text{cons}}$ alone inadvertently induces inter-head entropy variance, allowing $\mathcal{S}_{\text{var}}$ to maintain full detection with near-perfect ASR preserved. Under joint regularization, BadCLIP's ASR collapses to $14.93\%$, while BadViT maintains ASR at the cost of severe CA degradation ($40.66\%$), reflecting the irreconcilable conflict between evading both signals and preserving model utility. TrojViT results are identical across all settings, since the backdoor is injected into the classification head without modifying the encoder, confirming that our signals generalize to any trigger that influences attention, like adversarial examples. For BadVision, single regularization reduces ASR to $35.00\%$; joint regularization further degrades it to $11.00\%$. These results confirm that the proposed signals are unavoidable fingerprints of ViT backdoors, which cannot be suppressed without simultaneously destroying the attack.
 

\section{Conclusion}
This work probes the effect of backdoor triggers on the multi-head attention of ViTs and identifies two anomalous behaviors: complete and partial attention hijacking. Then, we propose \textbf{DetectViT}, a test-time backdoor detection method that quantifies these behaviors via an inter-head consistency score and an inter-head entropy variance score, with a small set of OOD calibration samples. Extensive experiments across various representative attacks, tasks, and models demonstrate that DetectViT outperforms all baselines, with negligible additional time overhead owing to its reliance solely on attention weights. Notably, DetectViT requires neither training data, model outputs, nor any prior knowledge of the trigger type, and applies to diverse ViT-based architectures and downstream tasks.

\paragraph{Limitations.}
While DetectViT demonstrates strong performance across diverse settings, several limitations remain. First, it cannot defend against OOD-as-trigger attacks~\cite{lyu2024backdooring}, where the backdoor is activated by natural OOD features without any crafted trigger, making poisoned samples indistinguishable from clean ones at inference time. Such challenges are shared by existing test-time detection methods. Second, the detection signals rely on CLS-to-patch attention weights, which may not generalize to architectures without a CLS token, leaving extension to such settings as future work.

\section*{Acknowledgments and Disclosure of Funding}
This work is supported by the National Natural Science Foundation of China (Grant No.~62476095),
the Guangdong Basic and Applied Basic Research Foundation (Grant No.~2025A1515011525), and the
Major Science and Technology Project of Yunnan Province (Grant No.~202502AD080017).
The authors declare no competing interests.

\bibliographystyle{plain}
\bibliography{main}

\appendix
\section*{Appendix}
\section{Main Algorithm}
\label{appendix:algorithm}
The full detection procedure is summarized in Algorithm~\ref{alg:detection}.

\begin{algorithm}[h!]
\caption{DetectViT}
\label{alg:detection}
\begin{algorithmic}[1]
\Require ViT encoder $f$, calibration set $\mathcal{D}_{\text{cal}} = \{x_m\}_{m=1}^{M}$, test sample $x$
\Ensure Prediction $\hat{y} \in \{0, 1\}$

\State \textbf{\textcolor{DodgerBlue4}{// Phase 1: Threshold Estimation}}
\For{each $x_m \in \mathcal{D}_{\text{cal}}$}
    \State Extract attention weights $\{\mathbf{A}^{(\ell,h)}\}$ from $f(x_m)$
    \State Compute $\tilde{\mathbf{a}}^{(\ell,h)}$ via \eqref{eq:normalize} for all $\ell, h$
    \State Compute $\mathcal{S}_{\text{cons}}(x_m)$ via \eqref{eq:cosine}--\eqref{eq:score_cons} and $\mathcal{S}_{\text{var}}(x_m)$ via \eqref{eq:entropy}--\eqref{eq:score_var}
\EndFor
\State Estimate $\mu_{\text{cons}}, \sigma_{\text{cons}}, \mu_{\text{var}}, \sigma_{\text{var}}$ via \eqref{eq:stats_cons}--\eqref{eq:stats_var}
\State Set thresholds $\tau_{\text{cons}}, \tau_{\text{var}}$ via \eqref{eq:threshold}

\State \textbf{\textcolor{DodgerBlue4}{// Phase 2: Test Sample Detection}}
\State Extract attention weights $\{\mathbf{A}^{(\ell,h)}\}$ from $f(x)$
\State Compute $\mathcal{S}_{\text{cons}}(x)$ via \eqref{eq:cosine}--\eqref{eq:score_cons} and $\mathcal{S}_{\text{var}}(x)$ via \eqref{eq:entropy}--\eqref{eq:score_var}
\If{$\mathcal{S}_{\text{cons}}(x) > \tau_{\text{cons}}$ \textbf{or} $\mathcal{S}_{\text{var}}(x) > \tau_{\text{var}}$}
    \State \Return $\hat{y} = 1$ \textcolor{DarkOliveGreen4}{\Comment{backdoor sample}}
\Else
    \State \Return $\hat{y} = 0$ \textcolor{DarkOliveGreen4}{\Comment{clean sample}}
\EndIf
\end{algorithmic}
\end{algorithm}

\section{Detailed Description of Experimental Setup}
\label{appendix:attack_details}
\begin{figure}[t]
	\centering
	\subfloat[BadViT]{\label{backdoor_fig:a}\includegraphics[width=0.22\textwidth]{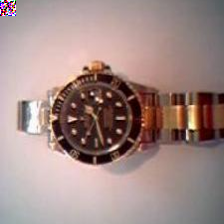}}\hspace{4mm}
	\subfloat[TrojViT]{\label{backdoor_fig:b}\includegraphics[width=0.22\textwidth]{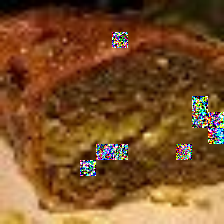}}\hspace{4mm}
    \subfloat[BadCLIP]{\label{backdoor_fig:c}\includegraphics[width=0.22\textwidth]{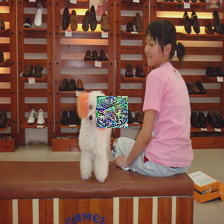}}\hspace{4mm}
    \subfloat[BadVision]{\label{backdoor_fig:d}\includegraphics[width=0.22\textwidth]{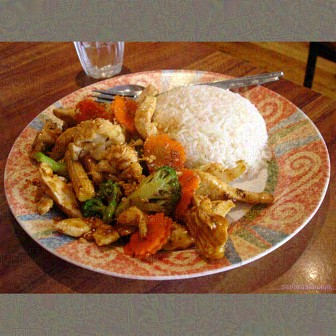}}
	\caption{Visualization of backdoor samples for each attack.}
    \label{backdoor_samples}
\end{figure}

\subsection{Attack Details}
We describe the attacks in our experiment in detail as follows:

\begin{itemize}
\item \textbf{BadViT \cite{yuan2023you}:} This model fine-tunes a pretrained DeiT-T vision encoder~\cite{pmlr-v139-touvron21a} on the Caltech-101 dataset while injecting the backdoor. The trigger is an optimized patch located in the top-left corner of the image with a size the same as a patch, as shown in Fig. \ref{backdoor_fig:a}. The original image size is $224 \times 224$. We set the target label index to $0$. Other training parameters are set to match those in the original paper.

\item \textbf{TrojViT \cite{zheng2023trojvit}:} This model fine-tunes a pretrained DeiT-S vision encoder~\cite{pmlr-v139-touvron21a} on the Tiny-ImageNet dataset while injecting the backdoor into the attention-related weights of the classification head rather than the encoder. Unlike fixed-position triggers, TrojViT employs multiple optimized patches whose positions are adaptively determined for each input image based on the attention map, as shown in Fig. \ref{backdoor_fig:b}. The original image size is $224 \times 224$. We set the target label index to $2$. Other training parameters are set to match those in the original paper.

\item \textbf{BadCLIP \cite{liang2024badclip}:} This attack targets the CLIP~\cite{radford2021learning} model (ViT-B-32 as the vision encoder) and pretrains it while injecting the backdoor on the CC3M dataset. The trigger is an optimized patch located in the center of the image with a size of $32 \times 32$, as shown in Fig. \ref{backdoor_fig:c}. The original image size in the CC3M dataset is $224 \times 224$. The reference label for the backdoor is ``banana''. Other training parameters are set to match those in the original paper.

\item \textbf{BadVision \cite{liu2025stealthy}:} This attack targets the vision encoder of LLaVA-1.5~\cite{liu2024improved} (clip-vit-large-patch14-336 as the vision backbone) and pretrains it while injecting the backdoor on the CGD subset of MIMIC-IT. The backdoor trigger is an optimized blended-based pattern with a size of $336 \times 336$, covering the entire image, as shown in Fig. \ref{backdoor_fig:d}. The reference concept for the backdoor is ``plane'', and the attack is considered successful if the generated caption contains this concept, e.g., \textit{The image features a large airplane parked on a runway, with its nose pointed towards the sky}. Other training parameters are set to match those in the original paper.
\end{itemize}

\subsection{Test Data Details}
We describe the test data in our experiment in detail as follows:
\begin{itemize}
\item \textbf{Caltech-101 \cite{li_andreeto_ranzato_perona_2022}:} The Caltech-101 dataset consists of images belonging to $101$ object categories plus a background category, with each category containing between $40$ and $800$ images. Images vary in size but are typically around $300 \times 200$ pixels, and are resized to $224 \times 224 \times 3$ during preprocessing. The dataset contains $9,144$ images in total, and is widely used for evaluating image classification models. After poisoning $5\%$ of the data, the number of backdoor images is $458$, and the number of clean images is $8,686$.

\item \textbf{Tiny-ImageNet \cite{le2015tiny}:} The Tiny-ImageNet dataset is a subset of ImageNet, consisting of images labeled with one of $200$ object categories. Each image is of size $64 \times 64 \times 3$ and is resized to $224 \times 224 \times 3$ during preprocessing. The dataset contains $100,000$ training images and $10,000$ validation images, and is widely used for evaluating lightweight image classification models. After poisoning $5\%$ of the data, the number of backdoor images is $500$, and the number of clean images is $9,500$.

\item \textbf{ImageNet 1K \cite{deng2009imagenet}:} The ImageNet 1K dataset consists of images labeled with one of $1,000$ object categories. Each image is of size $256 \times 256 \times 3$ and contains a single object from the corresponding category. The dataset contains $1.2$ million training images and $50,000$ validation images, and is widely used for evaluating image classification models. After poisoning $5\%$ of the data, the number of backdoor images is $2,500$, and the number of clean images is $47,500$.

\item \textbf{COCO Caption \cite{chen2015microsoft}:} The COCO Caption dataset is part of the larger COCO dataset, designed for image captioning and other vision-related tasks. It contains over $330,000$ images, with more than $200,000$ annotated with multiple captions describing the image's content. For our experiments, we randomly select $2,000$ samples to construct the test dataset, following the methodology of \cite{liu2025stealthy}. After poisoning $5\%$ of the data, the number of backdoor images is $100$, and the number of clean images is $1,900$.
\end{itemize}

\section{Additional Experiments}
\subsection{Validity of the Gaussian Assumption}
\label{app:gaussian}
 
The three-sigma rule in Eq.~(11) assumes that the detection scores of clean samples are approximately Gaussian. We examine this assumption on the clean test samples of each attacked model, using the inter-head consistency score $\mathcal{S}_{\text{cons}}$ for the three patch-based attacks (BadViT, TrojViT, BadCLIP) and the inter-head entropy variance score $\mathcal{S}_{\text{var}}$ for the blended-based attack (BadVision), consistent with Appendix~C.4.
 
We report two complementary measures in Table~\ref{tab:gaussian}. The first is the Shapiro-Wilk statistic $W$, which equals 1 for a perfectly Gaussian sample. The second is the fraction of samples falling within one, two, and three standard deviations of the mean, which we compare against the theoretical Gaussian values of 68.27\%, 95.45\%, and 99.73\%.
 
\begin{table}[h]
\centering
\caption{Gaussianity of the clean detection scores. We report the fraction of clean samples (\%) within one, two, and three standard deviations of the mean, together with the Shapiro-Wilk statistic $W$. Deviations from the theoretical Gaussian values are given in parentheses.}
\label{tab:gaussian}
\begin{tabular}{l cccc}
\toprule
 & 1-sigma & 2-sigma & 3-sigma & Shapiro-Wilk $W$ \\
\midrule
Theoretical & 68.27 & 95.45 & 99.73 & 1.0000 \\
\midrule
BadViT    & 69.80 ($+$1.53) & 94.65 ($-$0.80) & 99.79 ($+$0.06) & 0.9890 ($-$0.0110) \\
TrojViT   & 77.71 ($+$9.44) & 94.66 ($-$0.79) & 99.09 ($-$0.64) & 0.9179 ($-$0.0821) \\
BadCLIP   & 69.06 ($+$0.79) & 95.21 ($-$0.24) & 99.61 ($-$0.12) & 0.9973 ($-$0.0027) \\
BadVision & 67.47 ($-$0.80) & 95.32 ($-$0.13) & 99.68 ($-$0.05) & 0.9968 ($-$0.0032) \\
\bottomrule
\end{tabular}
\end{table}
 
On BadViT, BadCLIP, and BadVision, the clean scores are close to Gaussian: $W$ is near 1, and the coverage at all three levels deviates from theory by less than 1.6 percentage points. TrojViT departs the most ($W = 0.9179$), but the departure is concentrated at the peak rather than in the tails: 77.71\% of its clean scores fall within one standard deviation against a theoretical 68.27\%, meaning that its distribution is more sharply concentrated around the mean than a Gaussian, which favors detection. Its two- and three-sigma coverage remains close to theory, so the tail behavior on which the three-sigma threshold depends is unaffected.
 
To examine whether the detection performance hinges on the Gaussian assumption itself, we replace the three-sigma threshold with a distribution-free alternative, namely the 99th percentile of the OOD calibration scores, keeping all other settings unchanged. As the threshold does not affect the AUC, we report only TPR and FPR in Table~\ref{tab:percentile}.
 
\begin{table}[h]
\centering
\caption{Detection performance under the three-sigma threshold and the 99th-percentile threshold. TPR and FPR are reported as percentages (\%). We bold the best value.}
\label{tab:percentile}
\begin{tabular}{l cc cc cc cc}
\toprule
\multirow{2}{*}{Threshold}
 & \multicolumn{2}{c}{BadViT}
 & \multicolumn{2}{c}{TrojViT}
 & \multicolumn{2}{c}{BadCLIP}
 & \multicolumn{2}{c}{BadVision} \\
\cmidrule(lr){2-3} \cmidrule(lr){4-5} \cmidrule(lr){6-7} \cmidrule(lr){8-9}
 & TPR $\uparrow$ & FPR $\downarrow$
 & TPR $\uparrow$ & FPR $\downarrow$
 & TPR $\uparrow$ & FPR $\downarrow$
 & TPR $\uparrow$ & FPR $\downarrow$ \\
\midrule
Three-sigma     & \textbf{100.00} & \textbf{1.58} & \textbf{90.80} & 5.94 & \textbf{100.00} & \textbf{0.17} & 96.00 & \textbf{3.00} \\
99th percentile & \textbf{100.00} & 4.29 & 89.40 & \textbf{5.14} & \textbf{100.00} & 0.52 & \textbf{97.00} & 6.32 \\
\bottomrule
\end{tabular}
\end{table}
 
The two thresholds yield comparable detection performance across all four attacks, indicating that the effectiveness of DetectViT stems from the separability of the two signals rather than from the particular estimator used to derive the threshold. The three-sigma rule nonetheless achieves a lower FPR on three of the four attacks, which supports its use as the default.

\subsection{Evaluation on Additional Attacks}
\label{app:more_attacks}
 
To examine whether the two detection signals generalize beyond the attacks in the main evaluation, we consider five additional attacks spanning different threat models and trigger types: BadEncoder~\cite{jia2022badencoder} and DRUPE~\cite{tao2024distribution}, which implant backdoors into self-supervised visual encoders; Label-Consistent~\cite{turner2019label}, a clean-label attack; Low-Frequency~\cite{zeng2021rethinking}, which places the trigger in the frequency domain; and Blended-Optimized, which replaces the fixed pattern of the blended attack~\cite{chen2017targeted} with an optimized global trigger.
 
For BadEncoder and DRUPE, we follow the ViT protocol specified in the appendix of DRUPE~\cite{tao2024distribution}: we use Mugs~\cite{zhou2022mugs} as the self-supervised learning algorithm, pretrain ViT-T on CIFAR-10, and use GTSRB as the downstream task. For Label-Consistent and Low-Frequency, we reproduce the attacks from the implementations collected in BackdoorBench~\cite{wu2022backdoorbench}, training and testing ViT-S on Tiny-ImageNet. For Blended-Optimized, we replace the ``kitty'' pattern with an optimized global trigger and train within the BadCLIP framework (CLIP ViT-B/32), evaluating on ImageNet-1K. The detection settings (iNaturalist, $M=128$) and the baseline implementations are identical to those in the main evaluation.
 
\begin{table}[th]
\centering
\caption{Detection performance on five additional attacks. TPR and FPR are reported as percentages (\%). AUC is reported on a $[0,1]$ scale. We bold the best value.}
\label{tab:more_attacks}
\resizebox{\textwidth}{!}{
\begin{tabular}{l ccc ccc ccc ccc ccc}
\toprule
\multirow{2}{*}{Defense}
 & \multicolumn{3}{c}{BadEncoder}
 & \multicolumn{3}{c}{DRUPE}
 & \multicolumn{3}{c}{Low-Frequency}
 & \multicolumn{3}{c}{Label-Consistent}
 & \multicolumn{3}{c}{Blended-Optimized} \\
\cmidrule(lr){2-4} \cmidrule(lr){5-7} \cmidrule(lr){8-10} \cmidrule(lr){11-13} \cmidrule(lr){14-16}
 & TPR $\uparrow$ & FPR $\downarrow$ & AUC $\uparrow$
 & TPR $\uparrow$ & FPR $\downarrow$ & AUC $\uparrow$
 & TPR $\uparrow$ & FPR $\downarrow$ & AUC $\uparrow$
 & TPR $\uparrow$ & FPR $\downarrow$ & AUC $\uparrow$
 & TPR $\uparrow$ & FPR $\downarrow$ & AUC $\uparrow$ \\
\midrule
TeCo      & 54.91 & 62.29 & 0.37 & 48.10 & 65.79 & 0.35 & 99.60 & 62.38 & 0.52 & 66.80 & 56.99 & 0.57 & 12.12 & 29.67 & 0.42 \\
Scale-up  & 63.29 & 10.26 & 0.79 & 47.15 & 6.93  & 0.72 & \textbf{100.00} & 11.32 & 0.95 & 84.00 & 13.06 & 0.88 & 95.24 & 17.39 & 0.91 \\
PatchProc & 85.44 & 21.58 & 0.83 & 75.95 & 23.68 & 0.78 & \textbf{100.00} & 17.05 & 0.87 & 0.00  & 20.21 & 0.47 & 88.20 & 20.86 & 0.85 \\
DeDe      & 99.21 & 23.13 & 0.97 & \textbf{97.94} & 27.37 & 0.91 & 0.00 & 30.33 & 0.00 & 78.60 & 25.52 & 0.83 & 73.44 & 23.98 & 0.81 \\
B-IDS     & 96.04 & 17.12 & -    & 95.09 & 22.11 & -    & 0.00  & 26.00 & -    & 14.60 & 25.47 & -    & 44.96 & 12.78 & -    \\
\midrule
DetectViT & \textbf{99.37} & \textbf{5.33} & \textbf{0.99}
          & 95.57 & \textbf{4.42} & \textbf{0.99}
          & 93.40 & \textbf{7.11} & \textbf{0.97}
          & \textbf{94.00} & \textbf{6.52} & \textbf{0.97}
          & \textbf{99.08} & \textbf{0.64} & \textbf{1.00} \\
\bottomrule
\end{tabular}}
\end{table}
 
As shown in Table~\ref{tab:more_attacks}, DetectViT attains the best value in most cells and is the only method that maintains an AUC of at least 0.97 across all five attacks, whereas each baseline degrades severely on at least one of them. Label-based methods are sensitive to the trigger type: PatchProc fails to detect Label-Consistent (TPR 0.00\%), and TeCo reaches an AUC of only 0.42 on Blended-Optimized. Embedding-based methods fail on Low-Frequency, where both DeDe and BackdoorIDS detect none of the backdoor samples, suggesting that frequency-domain triggers leave the encoder representation largely indistinguishable from that of clean inputs.
 
DRUPE provides a notable case. Although it uses a patch trigger, its distribution-preserving objective suppresses inter-head consistency in pursuit of stealth and raises the inter-head entropy variance instead, so it is detected almost entirely by $\mathcal{S}_{\text{var}}$ rather than $\mathcal{S}_{\text{cons}}$. This mirrors the behavior observed under adaptive attacks in Table~5, where suppressing one signal induces the other. It also indicates that the visual form of a trigger does not reliably determine which signal is triggered; the two signals are complementary, and it is their disjunctive combination that makes the detection robust across trigger types.



\begin{figure}[t]
    \centering
    \includegraphics[width=1\textwidth]{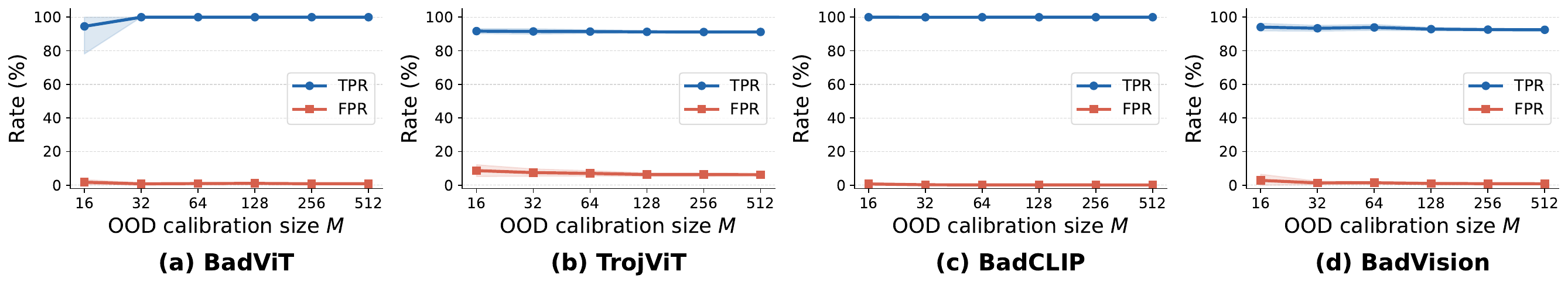}
    \caption{Effectiveness of DetectViT under different numbers of OOD calibration samples $M$.}
    \label{fig:complete_ood_size}
\end{figure}

\begin{figure}[t]
    \centering
    \includegraphics[width=1\textwidth]{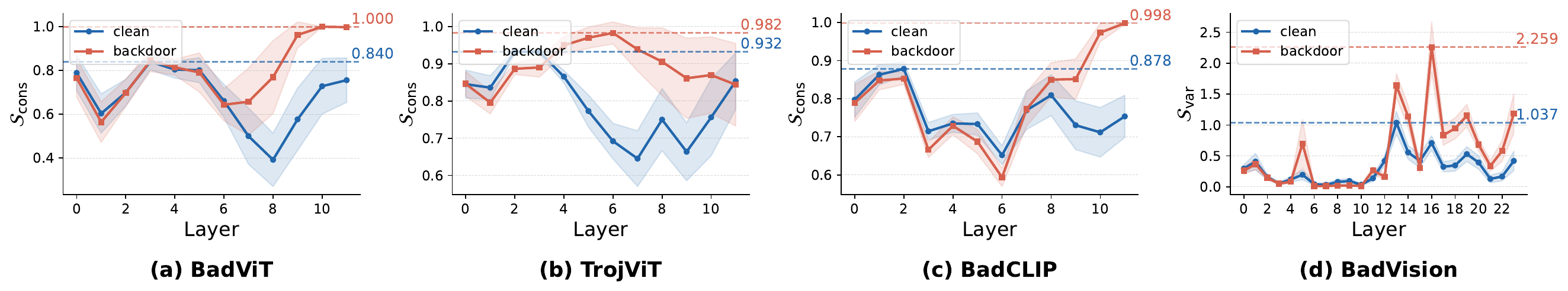}
    \caption{Mean value of detection signals across all backdoor/clean samples.}
    \label{fig:complete_per_layer_signals}
\end{figure}

\subsection{Detection Performance varying OOD Calibration Set Size}
\label{appendix:ood_size}
Fig.~\ref{fig:complete_ood_size} illustrates how detection performance varies with the number of OOD calibration samples $M$ across all four attacks. Each experiment is repeated under $10$ different random seeds, where the solid lines represent the mean value and the shaded regions indicate the standard deviation. Two consistent observations emerge. First, even with as few as $M=16$ samples, DetectViT already achieves strong detection performance across all attacks, demonstrating that the proposed threshold estimation is effective even in extremely data-scarce scenarios. Second, the standard deviation shrinks rapidly as $M$ increases: at $M=16$, the shaded region is noticeable on BadViT and BadVision, whereas at $M=32$ or $64$, the variance is negligible across all attacks, indicating that the thresholds have largely stabilized. This rapid convergence confirms that our method requires only a small number of OOD samples to produce reliable and stable thresholds. Based on these results, we adopt $M=128$ as the default setting, which provides a favorable balance between stability and calibration cost.

\subsection{Visualization of Detection Signals}
\label{appendix:visualization_signals}
Fig.~\ref{fig:complete_per_layer_signals} visualizes the per-layer detection signals averaged over all clean and backdoor samples across four attacks, where the solid lines and shaded regions represent the mean and standard deviation, respectively. For the three patch-based attacks (BadViT, TrojViT, BadCLIP), we report the inter-head cosine consistency $\mathcal{S}_{\text{cons}}$; for the blended-based attack BadVision, we report the inter-head entropy variance $\mathcal{S}_{\text{var}}$. In all cases, the peak signal value of backdoor samples substantially exceeds that of clean samples --- $1.000 $ vs. $ 0.840$ on BadViT, $0.982 $ vs. $ 0.932$ on TrojViT, $0.998 $ vs. $ 0.878$ on BadCLIP, and $2.259 $ vs. $ 1.037$ on BadVision --- confirming that the two proposed signals provide a reliable and discriminative separation between backdoor and clean inputs. Moreover, the peak signal tends to occur in the deeper layers of the model, suggesting that backdoor features are more prominently represented in posterior transformer layers.

\section{Code Availability}
We provide the implementation and detailed guidelines to reproduce our work in the code repository: \url{https://github.com/siquanhuang/DetectViT}.

\newpage
\input{checklist.tex}

\end{document}

%% file: checklist.tex
\section*{NeurIPS Paper Checklist}
\begin{enumerate}

\item {\bf Claims}
    \item[] Question: Do the main claims made in the abstract and introduction accurately reflect the paper's contributions and scope?
    \item[] Answer: \answerYes{} 
    \item[] Justification: The abstract and introduction clearly state our contributions.

\item {\bf Limitations}
    \item[] Question: Does the paper discuss the limitations of the work performed by the authors?
    \item[] Answer: \answerYes{} 
    \item[] Justification:  We explicitly discuss the limitations of our work in the section of conclusion.

\item {\bf Theory assumptions and proofs}
    \item[] Question: For each theoretical result, does the paper provide the full set of assumptions and a complete (and correct) proof?
    \item[] Answer: \answerNA{} 
    \item[] Justification: This paper is empirical in nature and does not include formal theorems or proofs. 

    \item {\bf Experimental result reproducibility}
    \item[] Question: Does the paper fully disclose all the information needed to reproduce the main experimental results of the paper to the extent that it affects the main claims and/or conclusions of the paper (regardless of whether the code and data are provided or not)?
    \item[] Answer: \answerYes{} 
    \item[] Justification: Sec. 5 and Appendix B provide detailed descriptions of the algorithm, attacked models, datasets, baselines, hyperparameters, and evaluation metrics. The full algorithm is given in Appendix A, and code is released at \url{https://github.com/siquanhuang/DetectViT}.

\item {\bf Open access to data and code}
    \item[] Question: Does the paper provide open access to the data and code, with sufficient instructions to faithfully reproduce the main experimental results, as described in supplemental material?
    \item[] Answer: \answerYes{} 
    \item[] Justification: We release code with reproduction instructions at \url{https://github.com/siquanhuang/DetectViT} (Sec. 5 and Appendix D). All datasets used (Caltech-101, Tiny-ImageNet, ImageNet-1K, COCO Caption, and the OOD calibration datasets) are publicly available.

\item {\bf Experimental setting/details}
    \item[] Question: Does the paper specify all the training and test details (e.g., data splits, hyperparameters, how they were chosen, type of optimizer) necessary to understand the results?
    \item[] Answer: \answerYes{} 
    \item[] Justification:  Sec. 5 specifies attacked models, datasets, baselines, evaluation metrics, hyperparameters (default OOD source iNaturalist with M=128), and hardware (single NVIDIA RTX 4090). Appendix B provides further details on attacks, trigger configurations, target labels, and test data splits.

\item {\bf Experiment statistical significance}
    \item[] Question: Does the paper report error bars suitably and correctly defined or other appropriate information about the statistical significance of the experiments?
    \item[] Answer: \answerYes{} 
    \item[] Justification: For experiments with random calibration sampling (Fig. 3 and Fig. 6), each setting is repeated under 10 random seeds, and we report the mean values along with shaded regions indicating the standard deviation. Per-layer detection signal visualizations (Fig. 4 and Fig. 7) likewise report mean and standard deviation across all samples.

\item {\bf Experiments compute resources}
    \item[] Question: For each experiment, does the paper provide sufficient information on the computer resources (type of compute workers, memory, time of execution) needed to reproduce the experiments?
    \item[] Answer: \answerYes{} 
    \item[] Justification: Sec. 5 states that all experiments are conducted on a single NVIDIA GeForce RTX 4090 GPU. Per-sample detection runtime for all methods is reported in Table 3, demonstrating that DetectViT introduces negligible overhead since it reuses attention weights already computed during inference.
    
\item {\bf Code of ethics}
    \item[] Question: Does the research conducted in the paper conform, in every respect, with the NeurIPS Code of Ethics \url{https://neurips.cc/public/EthicsGuidelines}?
    \item[] Answer: \answerYes{} 
    \item[] Justification: This work conforms with the NeurIPS Code of Ethics.

\item {\bf Broader impacts}
    \item[] Question: Does the paper discuss both potential positive societal impacts and negative societal impacts of the work performed?
    \item[] Answer: \answerYes{} 
    \item[] Justification: Our work has clear positive societal impact: it provides a practical defense against backdoor attacks on widely deployed Vision Transformers and multimodal models, helping practitioners safely use third-party pretrained checkpoints. As a defensive method, it does not introduce direct negative societal impacts; while a sophisticated adversary could in principle adapt to our signals, our adaptive-attack experiments (Table 5) show that evasion comes at the cost of either attack success or model utility.
    
\item {\bf Safeguards}
    \item[] Question: Does the paper describe safeguards that have been put in place for responsible release of data or models that have a high risk for misuse (e.g., pre-trained language models, image generators, or scraped datasets)?
    \item[] Answer: \answerNA{} 
    \item[] Justification:  This paper releases a backdoor detection method rather than pretrained generative models, scraped datasets, or other assets with high misuse risk. The released code is purely defensive and does not facilitate the creation of new attacks.

\item {\bf Licenses for existing assets}
    \item[] Question: Are the creators or original owners of assets (e.g., code, data, models), used in the paper, properly credited and are the license and terms of use explicitly mentioned and properly respected?
    \item[] Answer:  \answerYes{}
    \item[] Justification: All existing assets used in this work — including DeiT, CLIP, LLaVA-1.5, and the datasets (Caltech-101, Tiny-ImageNet, ImageNet-1K, COCO Caption, iNaturalist, Food101, Oxford-IIIT Pet, Places365, STL-10, Flowers102, PASCAL VOC) — are properly cited in the references, and we use them in accordance with their respective licenses and terms of use.

\item {\bf New assets}
    \item[] Question: Are new assets introduced in the paper well documented and is the documentation provided alongside the assets?
    \item[] Answer: \answerYes{} 
    \item[] Justification: We release the implementation of DetectViT as a code repository (\url{https://github.com/siquanhuang/DetectViT}) with documentation and instructions for reproducing all reported results, as described in Appendix D.

\item {\bf Crowdsourcing and research with human subjects}
    \item[] Question: For crowdsourcing experiments and research with human subjects, does the paper include the full text of instructions given to participants and screenshots, if applicable, as well as details about compensation (if any)? 
    \item[] Answer: \answerNA{} 
    \item[] Justification: This work does not involve crowdsourcing experiments or research with human subjects. All experiments are conducted on publicly available image and image-caption datasets.

\item {\bf Institutional review board (IRB) approvals or equivalent for research with human subjects}
    \item[] Question: Does the paper describe potential risks incurred by study participants, whether such risks were disclosed to the subjects, and whether Institutional Review Board (IRB) approvals (or an equivalent approval/review based on the requirements of your country or institution) were obtained?
    \item[] Answer: \answerNA{} 
    \item[] Justification: This work does not involve human subjects research and therefore does not require IRB approval.

\item {\bf Declaration of LLM usage}
    \item[] Question: Does the paper describe the usage of LLMs if it is an important, original, or non-standard component of the core methods in this research? Note that if the LLM is used only for writing, editing, or formatting purposes and does \emph{not} impact the core methodology, scientific rigor, or originality of the research, declaration is not required.
    \item[] Answer: \answerNA{} 
    \item[] Justification: LLMs were only used for language polishing and editing of the manuscript. 

\end{enumerate}